\documentclass[12pt]{cmuthesis}

\usepackage{times}
\usepackage{fullpage}
\usepackage{graphicx}
\usepackage{amsmath}
\usepackage[numbers,sort]{natbib}
\usepackage[backref,pageanchor=true,plainpages=false, pdfpagelabels, bookmarks,bookmarksnumbered,
]{hyperref}
\usepackage{subfigure}
\usepackage{color, colortbl}

\usepackage{xcolor}
\hypersetup{
	colorlinks,
	linkcolor={red!50!black},
	citecolor={blue!50!black},
	urlcolor={blue!80!black}
}

\usepackage[letterpaper,twoside,vscale=.8,hscale=.75,nomarginpar,hmarginratio=1:1]{geometry}

\begin{document} 
\frontmatter

\pagestyle{empty}

\title{{
{\bf Industrial Smoke Detection and Visualization}} \vspace{1em}}
\author{Yen-Chia Hsu, Paul Dille, Randy Sargent, Illah Nourbakhsh}
\date{September 2016}
\Year{2016}
\trnumber{CMU-RI-TR-16-55}

\support{}
\disclaimer{}



\maketitle


\pagestyle{plain} 


\begin{acknowledgments}
This work is supported by the Heinz Endowments. The authors thank the Allegheny County Clean Air Now (ACCAN) community for participating in using the system presented in this work. The authors also thank Aaron Steinfeld and Srinivasa Narasimhan in CMU Robotics for providing academic feedback.
\end{acknowledgments}

\tableofcontents

\newcommand{\minus}{\scalebox{1}[0.9]{\boldmath$-$}}
\newcommand{\abs}[1]{\left\lvert#1\right\rvert}
\newcommand{\bgSub}{\mathrm{bgSub}}
\newcommand{\nega}{\text{-}}
\newcommand{\posi}{\text{+}}
\newcommand{\argmax}{\operatornamewithlimits{argmax}}
\newcommand{\imSub}{\mathrm{imSub}}

\definecolor{gray}{rgb}{0.85,0.85,0.85}
\definecolor{red}{rgb}{1,0,0}
\definecolor{blue}{rgb}{0,0,1}
\newcommand{\redtxt}[1]{\textcolor{red}{#1}}
\newcommand{\bluetxt}[1]{\textcolor{blue}{#1}}

\mainmatter


\begin{abstract}
	
As sensing technology proliferates and becomes affordable to the general public, there is a growing trend in citizen science where scientists and volunteers form a strong partnership in conducting scientific research including problem finding, data collection, analysis, visualization, and storytelling. Providing easy-to-use computational tools to support citizen science has become an important issue. To raise the public awareness of environmental science and improve the air quality in local areas,  we are currently collaborating with a local community in monitoring and documenting fugitive emissions from a coke refinery. We have helped the community members build a live camera system which captures and visualizes high resolution timelapse imagery starting from November 2014. However, searching and documenting smoke emissions manually from all video frames requires manpower and takes an impractical investment of time. This paper describes a software tool which integrates four features: (1) an algorithm based on change detection and texture segmentation for identifying smoke emissions; (2) an interactive timeline visualization providing indicators for seeking to interesting events; (3) an autonomous fast-forwarding mode for skipping uninteresting timelapse frames; and (4) a collection of animated smoke images generated automatically according to the algorithm for documentation, presentation, storytelling, and sharing. With the help of this tool, citizen scientists can now focus on the content of the story instead of time-consuming and laborious works.

\end{abstract}

\chapter{Introduction}

As sensing devices become available and affordable to the general public, there is a growing trend in citizen science~\cite{Haklay2013} where non-professionals find and define problems, collect and analyze environmental data by using sensors in order to better understand their surroundings, and to tell compelling stories in communicating findings. Storytelling is an important technique in communication, since visual stories framed by data are easy to remember, encourage discussions, support decision making, and have the potential to raise public awareness~\cite{Kosara2013,Ma2012}. However, citizen scientists often lack related skills and require the assistance of experts in building sensor networks, analyzing large-scale environmental data, and integrating the findings into scientific stories. Therefore, developing and providing easy-to-use tools matching the needs of local communities is vital in helping them improve technology fluency~\cite{Silvertown2009}.

\begin{figure}
	\centering
	\includegraphics[width=1\columnwidth]{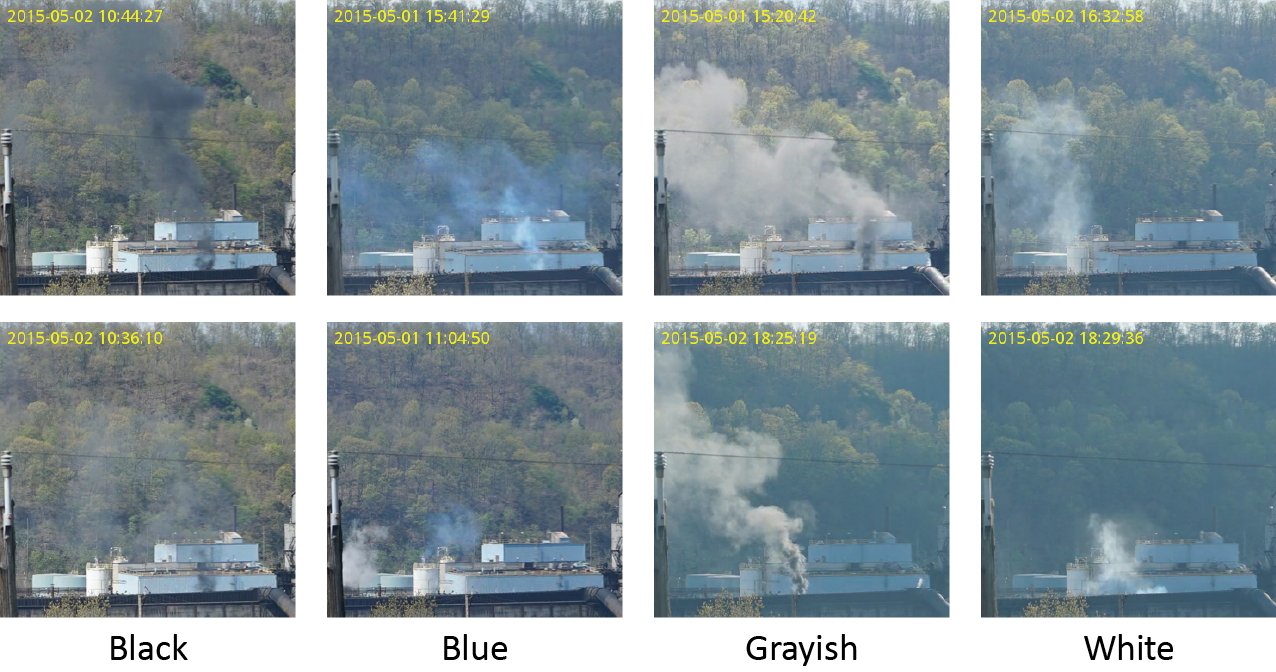}
	\caption{This figure shows emissions with various lightings, appearance, and opacities.}
	\label{fig:smoke-type-1}
\end{figure}

\begin{figure}
	\centering
	\includegraphics[width=1\columnwidth]{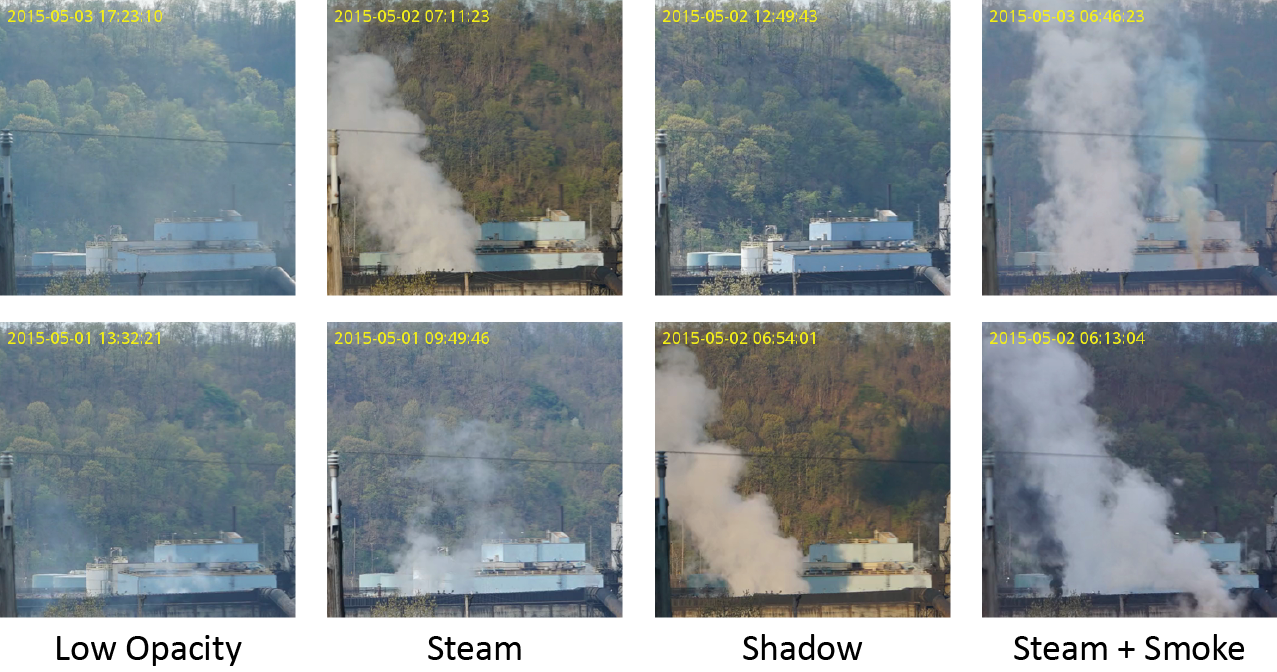}
	\caption{This figure shows steam, shadow, and the mixture of steam and smoke.}
	\label{fig:smoke-type-2}
\end{figure}

We are currently collaborating with a local community in Pittsburgh in gathering air quality data (particle pollution PM2.5), documenting images of fugitive emissions from a coke refinery, and presenting scientific stories about how smoke emissions affect the local air quality. Figure~\ref{fig:smoke-type-1} demonstrates smoke emissions with various lightings, appearance, and opacities. Figure~\ref{fig:smoke-type-2} shows steam, shadow, and the mixture of steam and smoke.

We have helped the local community build a live camera monitoring system. The system collects high resolution imagery starting from November 2014 and visualizes the results by using a web-based large-scale timelapse viewer which we developed previously~\cite{Sargent2010,TMviewer}. The local community can utilize the viewer to explore the high quality time-series images by panning and zooming, finding fugitive emissions, and using the provided thumbnail tool to generate and share animated images. Nevertheless, the speed of receiving images greatly exceeds the time required to process them. The live camera takes a picture every 5 seconds and a timelapse for one day contains nearly 17,000 frames. Manually searching through each image to identify smoke emissions takes more than six times as long without computer vision automation. This paper presents a computer vision tool for detecting industrial smoke emissions and generating related animated images automatically, which significantly reduces the workload of citizen scientists. The task is to detect frames from a static camera containing smoke, exclude the ones having steam and shadow, identify the starting and ending frames of emissions, and output animated images which may include smoke. The local community can easily select representative images and insert them directly into Google Docs as animated image sequences to form a collection of fugitive emissions.

\chapter{Related Work}

There are three general approaches appearing in previous research for detecting the presence of smoke emissions in a single image or across multiple frames: (1) color modeling; (2) change detection; and (3) texture analysis.

Color modeling describes the characteristics of image intensity values. For instance, smoke is grayish and has low saturation. Previous research used color models to identify smoke pixels~\cite{Celik2007} or extract color histogram features~\cite{Lee2012}.

Change detection~\cite{Radke2005} determines moving objects in an image, which provides candidate regions containing smoke emissions for further analysis. One common technique is background subtraction~\cite{Collins2000,Cheung2005} which estimates an image without moving objects from an image sequence, subtracts the estimated image from the current one to get a residual image, and thresholds the residual image to obtain a binary mask. In addition, there are background modeling approaches~\cite{Stauffer1999,Friedman1997} which learn a probabilistic model of each pixel using a mixture-of-Gaussians and determine the background pixels according to the probability distribution. Other techniques involve computing the entropy of the optical flow field~\cite{Kopilovic2000} to identify smoke and checking flickering pixels at the edge of candidate smoke regions~\cite{Toreyin2005}.

Texture analysis measures texture energy in a single image or texture changes between multiple frames. One approach is to apply texture descriptors, such as a wavelet transform, on small blocks in an image for obtaining feature vectors and train a classifier using these features~\cite{Gubbi2009,Calderara2008}.

Each of these approaches has distinct strengths and weaknesses. Color modeling is straightforward, but suffers from situations where smoke and non-smoke objects have the same chrominance (e.g. white smoke and steam, dark shadow and black smoke) or the background does not contain plentiful color information due to various weather and lighting conditions (e.g. fog, nighttime images). Background subtraction and background modeling do not distinguish smoke from non-smoke regions since they find all moving objects including shadow, steam, and smoke. Optical flow can determine smoke motions, but has high computational cost. It is difficult to extract useful information from texture analysis if the background does not contain sufficient texture information. Several research has integrated these methods into a system for better performance. Toreyin et al.~\cite{Toreyin2005} combined background subtraction, edge flickering, and texture analysis into a final result. Lee et al.~\cite{Lee2012} used change detection to extract candidate regions, computed feature vectors based on color modeling and texture analysis, and trained a support vector machine classifier using these features.

We are aware of other advanced machine learning approaches. For instance, Hohberg~\cite{Hohberg2015} trains a convolutional neural network for recognizing wildfire smoke. Tian et al.~\cite{Tian2015} present a physical based model and use sparse coding to extract reliable features for single image smoke detection. However, a simpler heuristic approach combining color modeling, change detection, and texture analysis is sufficient for our current needs.


\chapter{System}

This chapter describes the method of the smoke detection algorithm, the experiment for evaluating the performance, and the visualization for showing smoke detection results.

\section{Method}

Inspired by prior method integration approaches, we have implemented a smoke detection algorithm for detecting fugitive emissions during the daytime from a static camera. The algorithm contains five steps: preprocessing, change detection, texture segmentation, region filtering, and event detection. Change detection identifies moving pixels containing smoke, steam, and shadow. Texture segmentation clusters pixels into several candidate regions based on texture information. Region filtering iteratively evaluates each candidate region based on shape, color, size, and the amount of change to determine if it matches the appearance and behavior of smoke. Event Detection groups video frames with smoke together to identify the starting and ending time of fugitive emissions.

\subsection{Preprocessing}

We apply the algorithm on 9700 daytime frames for each day and ignore nighttime. To reduce the computational cost, we first scale the original image at time $t$ down to one-fourth of the original size to obtain a downsampled image $I_{t}$. Then we estimate the background image $B_{t}$ by taking the median over the previous 60 images as shown in (\ref{eq:median}).
\begin{equation}\label{eq:median}
B_{t}(x,y) = \mathrm{median}\big(I_{t}(x,y),...,I_{t-59}(x,y)\big)
\end{equation}
where $(x,y)$ indicates the position of a pixel. Finally we convert all RGB images with 8-bit unsigned integer format to double precision ranging from 0 to 1.

\subsection{Change Detection}

Change detection finds moving pixels in video frames by computing changes in high frequency signals (e.g. edges, textures) and image intensity values (e.g. colors).

\subsubsection{High Frequency Change Detection}\label{sec:HFCD}

\begin{figure}
	\centering
	\includegraphics[width=1\columnwidth]{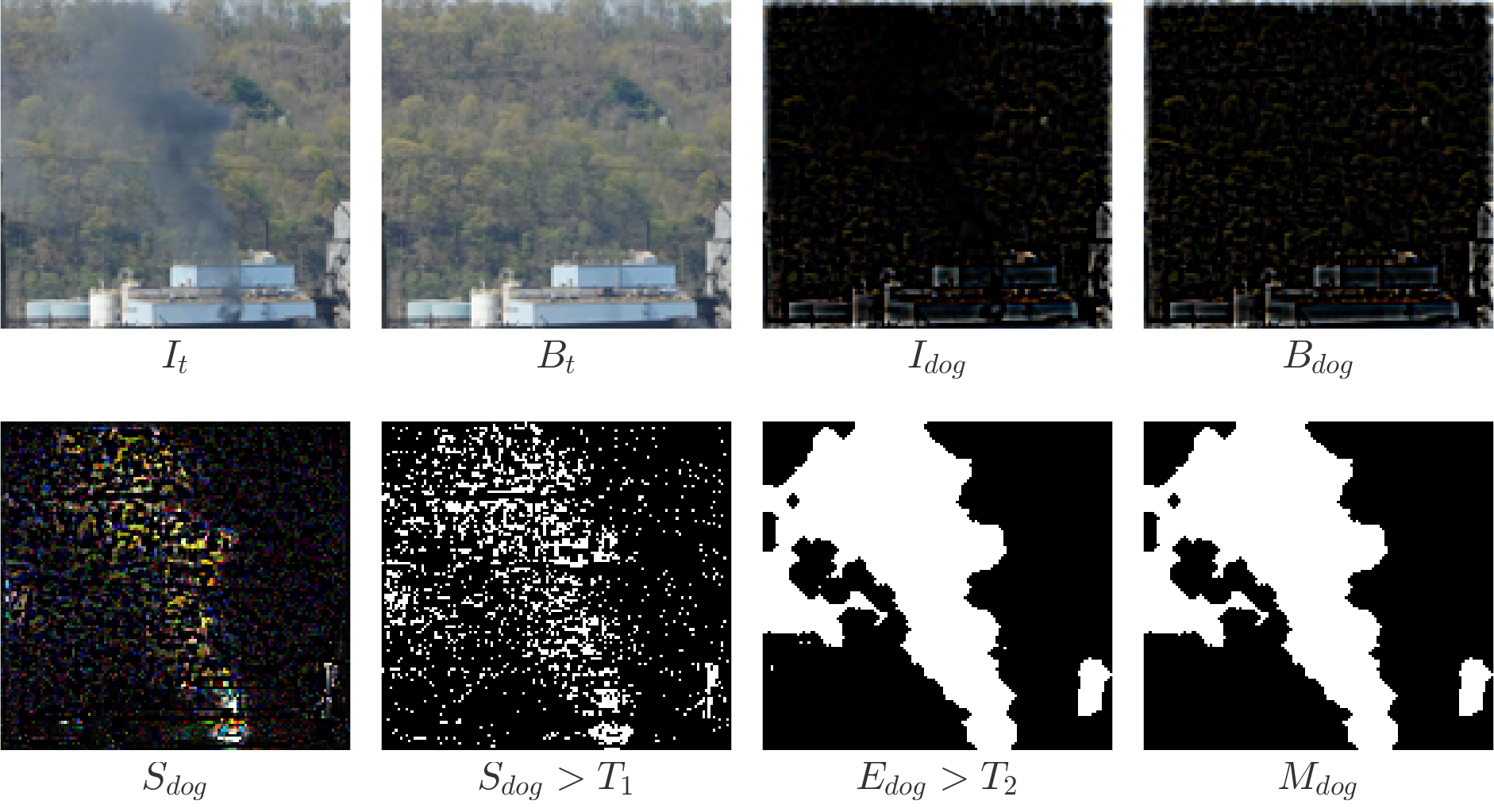}
	\caption{This figure visualizes the steps of high frequency change detection. Refer to section \ref{sec:HFCD} for detailed explanation.}
	\label{fig:HFCD}
\end{figure}

Smoke is semi-transparent with various opacities and occludes parts of the background upon presence, which causes changes of high frequency signals across frames. First we compute the difference of Gaussian (DoG) of $I_{t}$ and $B_{t}$ to obtain $I_{dog}$ and $B_{dog}$ as shown in (\ref{eq:DoG})
\begin{equation}
G_\sigma(x,y) = \frac{1}{2 \pi \sigma^2}\mathrm{exp}\Big(-\frac{x^2+y^2}{2 \sigma^2}\Big)
\end{equation}
\begin{equation}\label{eq:DoG}
\begin{array}{lll}
\! I_{dog}(x,y) & = & \big(G_{\sigma_{1}}(x,y)-G_{\sigma_{2}}(x,y)\big) \ast I_{t}(x,y) \\
\! B_{dog}(x,y) & = & \big(G_{\sigma_{1}}(x,y)-G_{\sigma_{2}}(x,y)\big) \ast B_{t}(x,y)
\end{array}
\end{equation}
where the asterisk sign $*$ indicates the convolution operator and $G_{\sigma}(x,y)$ is a Gaussian kernel with variance $\sigma^2$ and mean zero. The DoG image contains high frequency information for the current and the background images.

Then we perform background subtraction on $I_{dog}$ and $B_{dog}$ to obtain $S_{dog} = \bgSub(I_{dog},B_{dog})$ as shown in (\ref{eq:BS}).
\begin{equation}\label{eq:BS}
\bgSub(I,B) = \frac{\abs{I-B}}{\max\big(I+B,0.1\big)}
\end{equation}
Dividing the background subtraction term in the nominator by $\max\big(I+B,0.1\big)$ alleviates the effect of illumination in images. The $\max$ function in the denominator in (\ref{eq:BS}) prevents dividing to an extremely small value or zero. One way to interpret the $S_{dog}$ image is that it measures the change of high frequency signals such as edges and texture between the current and background image. Thresholding channels in $S_{dog}$ yields a binary image. Computing the local entropy of the 9-by-9 neighborhood centered around each pixel in the binary image gives an entropy image $E_{dog}$ as show in (\ref{eq:entropy}).
\begin{equation}\label{eq:entropy}
E_{dog} = \mathrm{entropyFilter}\big(\bgSub(I_{dog},B_{dog})>T_{1}\big)
\end{equation}

Finally we threshold the entropy image $E_{dog}$ to obtain a binary image $E_{dog}>T_{2}$. Performing morphological closing, removing noise using a median filter, and discarding small regions using connected component algorithm on the binary image yields the smoothed image $M_{dog}$ as shown in (\ref{eq:HFCD}).
\begin{equation}\label{eq:HFCD}
M_{dog} = \mathrm{smooth}(E_{dog}>T_{2})
\end{equation}
Figure~\ref{fig:HFCD} visualizes the steps of high frequency change detection. If the $M_{dog}$ image contains no regions (i.e. all pixel values are zero), the smoke detection algorithm terminates at this step and outputs zero as the response.

\subsubsection{Image Intensity Change Detection}\label{sec:IICD}

\begin{figure}
	\centering
	\includegraphics[width=1\columnwidth]{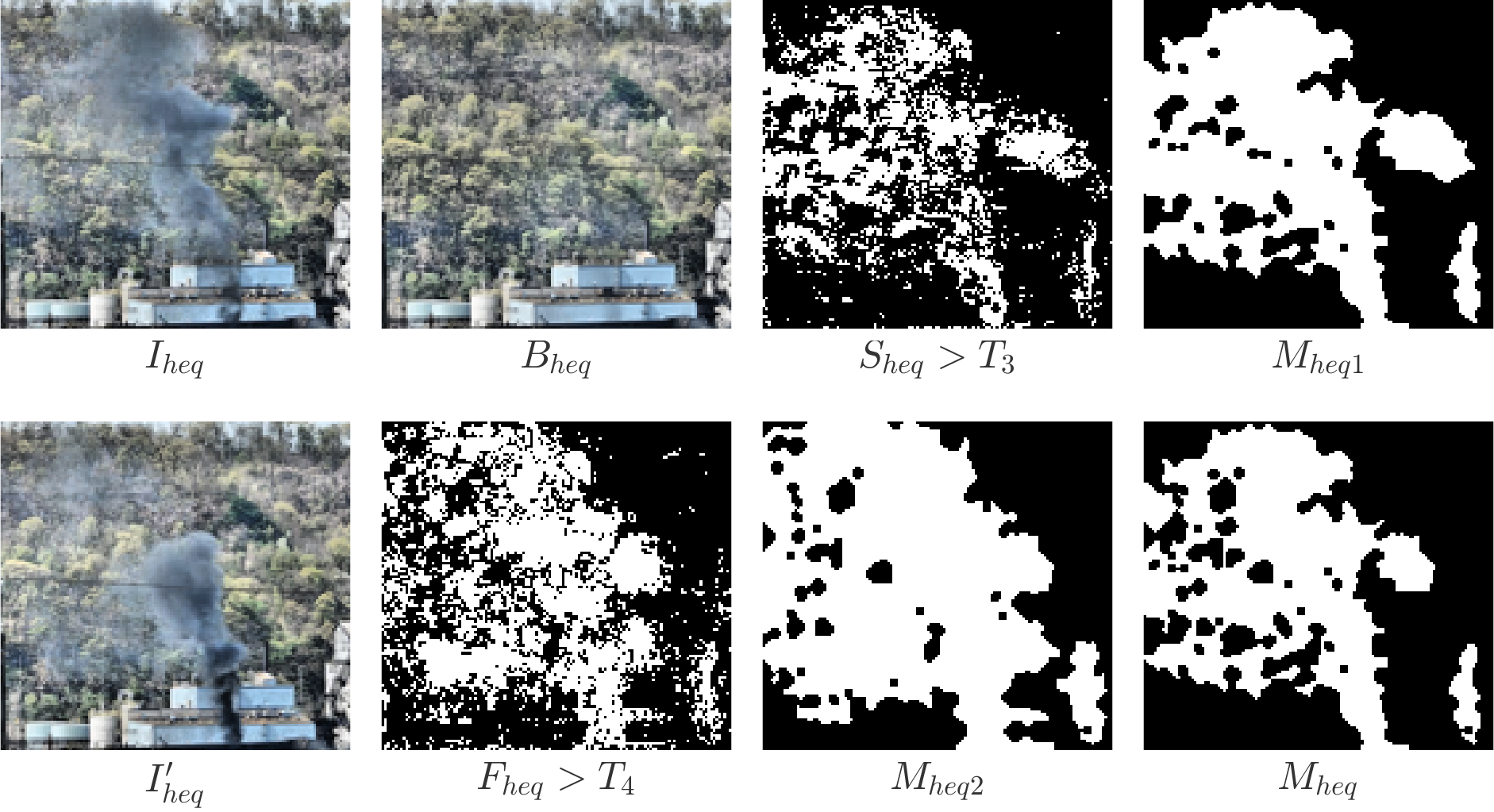}
	\caption{This figure visualizes the steps of image intensity change detection. Refer to section \ref{sec:IICD} for detailed explanation.}
	\label{fig:IICD}
\end{figure}

Changes of pixel intensity values across frames indicate candidate regions containing smoke. We first enhance the contrast of image $I_{t}$, $I_{t-2}$, and $B_{t}$ by using CLAHE (contrast-limited adaptive histogram equalization~\cite{Zuiderveld1994}) to obtain $I_{heq}$, $I'_{heq}$, and $B_{heq}$. CLAHE limits the contrast to avoid over-amplifying noise and operates on small local regions in the image. The desired shape of the histogram in a local region is approximately flat and follows a uniform distribution. One reason for performing contrast enhancement is that the color and saturation of smoke may be similar to the background under some lighting conditions.

Next we perform background subtraction as shown in (\ref{eq:BS}) on each channel of the two image pairs $(I_{heq},B_{heq})$ and $(I_{heq},I'_{heq})$ to obtain $S_{heq} = \bgSub(I_{heq},B_{heq})$ and $F_{heq} = \bgSub(I_{heq},I'_{heq})$, which provides information about the change of image intensity values between the current frame, background, and the previous frame. Smoothing the binary images $S_{heq}>T_{3}$ and $F_{heq}>T_{4}$ by using the process described in section \ref{sec:HFCD} yields $M_{heq1}$ and $M_{heq2}$.

Finally we combine $M_{heq1}$ and $M_{heq2}$ by using an AND operator into the resulting image $M_{heq}$ as shown in (\ref{eq:IICD}). Figure~\ref{fig:IICD} visualizes the steps of image intensity change detection.
\begin{equation}\label{eq:IICD}
\begin{array}{lll}
\! M_{heq1} & = & \mathrm{smooth}\big(\bgSub(I_{heq},B_{heq})>T_{3}\big) \\
\! M_{heq2} & = & \mathrm{smooth}\big(\bgSub(I_{heq},I'_{heq})>T_{4}\big) \\
\! M_{heq} & = & M_{heq1} \;\;\mathrm{and}\;\; M_{heq2}
\end{array}
\end{equation}

\subsection{Texture Segmentation}\label{sec:tex-seg}

\begin{figure}
	\centering
	\includegraphics[width=1\columnwidth]{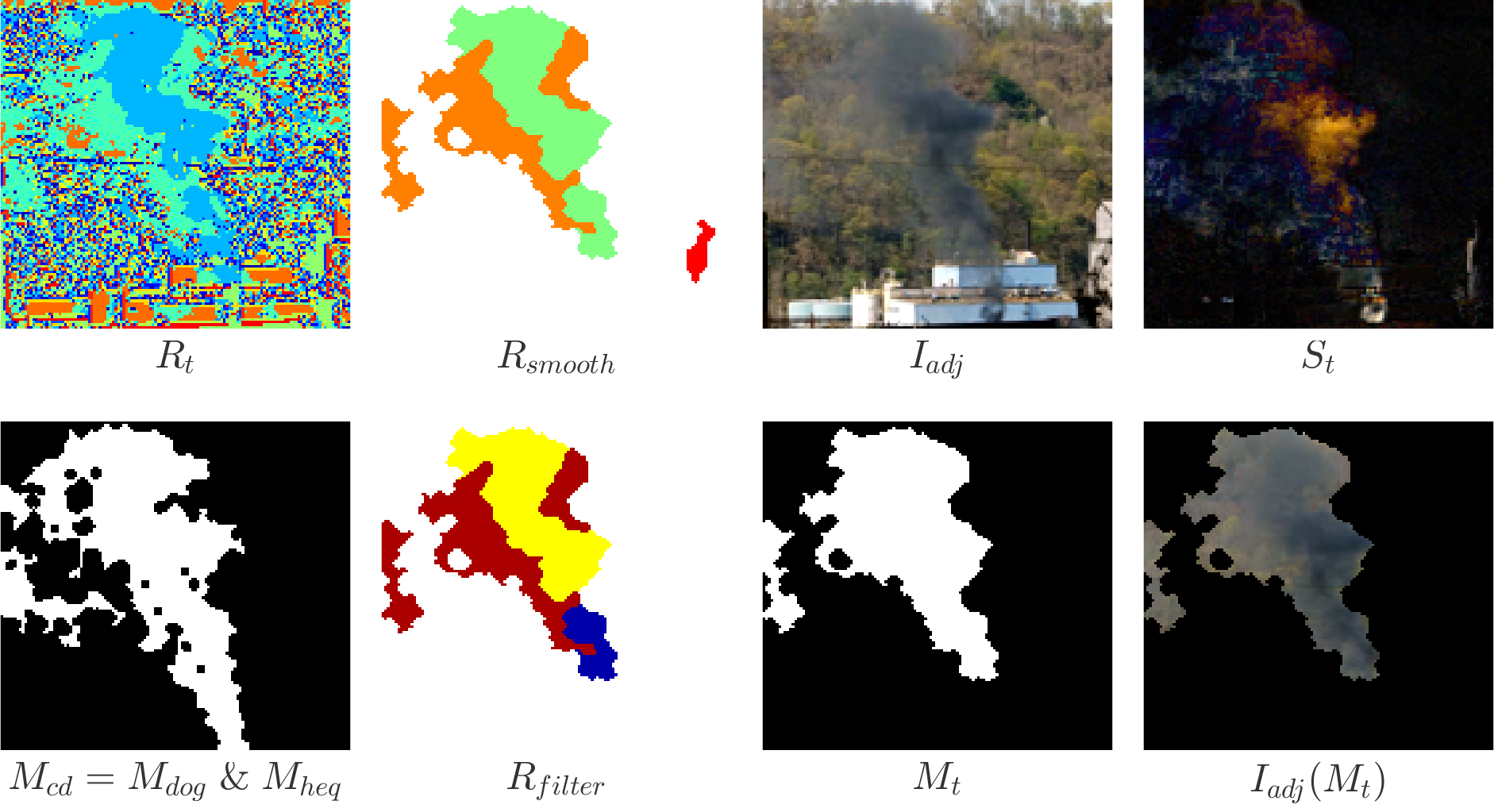}
	\caption{This figure demonstrates the steps of texture segmentation and region filtering. See section \ref{sec:tex-seg} and \ref{sec:region-filtering} for detailed explanation.}
	\label{fig:TEX}
\end{figure}

Texture segmentation partitions images into regions based on their texture information. This step computes filter responses by convolving an image with a filter bank, clusters the responses into a set of textons~\cite{Malik2001}, and partitions the image into separate regions by using these textons. We first combine the results of change detection algorithms by performing an AND operation on $M_{dog}$ and $M_{heq}$ to obtain $M_{cd}$ as shown in Figure~\ref{fig:TEX}. If all pixel values in image $M_{cd}$ are zero, the smoke detection algorithm stops at this step and outputs zero as the response.

Next we compute the filter bank using a variation of Laws' texture energy measures~\cite{Laws1980} as shown in (\ref{eq:Laws}).
\begin{equation}\label{eq:Laws}
\begin{array}{lrrrrrrrrl}
L5 & = & [ & 1 & 2 & 3 & 2 & 1 & ] & (\mathrm{Level})\\
E5 & = & [ & \nega 1 & \nega 2 & 0 & 2 & 1 & ] & (\mathrm{Edge})\\
S5 & = & [ & \nega 1 & 0 & 2 & 0 & \nega 1 & ] & (\mathrm{Spot})\\
W5 & = & [ & \nega 1 & 2 & 0 & \nega 2 & 1 & ] & (\mathrm{Wave})\\
R5 & = & [ & 1 & \nega 4 & 6 & \nega 4 & 1 & ] & (\mathrm{Ripple})
\end{array}
\end{equation}
The filter bank is a set of 5-by-5 convolution masks obtained by calculating the outer products of pairs of texture vectors in (\ref{eq:Laws}). The L5, E5, S5, W5, and R5 vectors detects gray level, edges, spots, waves, and ripples in the image respectively.

Then we take the contrast-enhanced image $I_{heq}$, subtract it with the mean value of $I_{heq}$, and convolve it with the filter bank for each RGB channel to obtain feature vectors. Each vector represents the corresponding pixel in $I_{heq}$ in the feature space and has 125 dimensions. Then the algorithm performs Principal Component Analysis which preserves 98\% of the energy (eigenvalues) on the feature vectors to reduce dimensions. Using the contrast-enhanced image alleviates the problem that some weather circumstances such as fog cause a decrease in background texture information.

Finally we perform an accelerated k-means++ algorithm~\cite{Arthur2007,Elkan2003} which chooses better initialized values (seed points) to cluster the feature vectors into textons and divide the current image into various regions as shown in image $R_{t}$ in Figure~\ref{fig:TEX}. Smoothing the image $R_{t}$ by discarding small regions, removing noise by using a median filter, and performing morphological closing yields $R_{smooth}$ in Figure~\ref{fig:TEX}.

\subsection{Region Filtering}\label{sec:region-filtering}

Region filtering determines if a region matches the appearance and behavior of smoke by evaluating shape, color, size, and the amount of change. We first use the connected component algorithm to find all separated regions and remove the ones which are thin and narrow. Mathematically speaking, for each region, the ratio of width to height of its bounding box exceeds a certain threshold. Or the ratio of the size of the region and its bounding box is smaller than a threshold.

Next we adjust the contrast of each channel in $I_{t}$ to produce $I_{adj}$ in Figure~\ref{fig:TEX} by stretching intensity values so that 1\% of the data is saturated at low and high intensities of $I_{t}$. We group nearby white regions and black ones based on $I_{adj}$ to reconstruct the shapes of objects. Since the color of smoke is usually grayish or bluish, we can remove regions having non-grayish and non-bluish colors described by (\ref{eq:non-grayish})
\begin{equation}\label{eq:non-grayish}
\begin{array}{lll}
\!\!\!\!\!\!\!\! \abs{c_{1} \minus c_{2}} \ge t_{1} \;\;\mathrm{or} \;\; \abs{c_{2} \minus c_{3}} \ge t_{2} \;\;\mathrm{or}\;\; \abs{c_{1} \minus c_{3}} \ge t_{3} \\
\!\!\!\!\!\!\!\! c_{j} = \mathrm{median}\big(I_{adj}(x,y,j)\big) \;\; \forall (x,y) \in R_{i}
\end{array}
\end{equation}
where $j$ indicates different channels in $I_{adj}$, $R_{i}$ denotes the $i^{th}$ region, $(x,y)$ means the location of pixels, and $\{c_{j}:j=1,2,3\}$ are the median of corresponding pixel values in $R_{i}$ in the RGB channels of $I_{adj}$. We also remove regions having light colors described by (\ref{eq:non-white}) because steam is usually white.
\begin{equation}\label{eq:non-white}
c_{1} \ge t_{4} \;\;\mathrm{and}\;\; c_{2} \ge t_{5} \;\;\mathrm{and}\;\; c_{3} \ge t_{6}
\end{equation}

Then we compute the size of each region and remove large or small ones which may be noise and shadow respectively. Furthermore, we remove the $i^{th}$ region $R_{i}$ if it does not have sufficient amount of change by summing up the corresponding pixel values in $M_{cd}$ by using (\ref{eq:change})
\begin{equation}\label{eq:change}
\sum_{\forall (x,y) \in R_{i}}{\!\!M_{cd}(x,y)} \le t_{7}
\end{equation}
where $(x,y)$ denotes the location of pixels in region $R_{i}$.

\begin{figure}
	\centering
	\includegraphics[width=1\columnwidth]{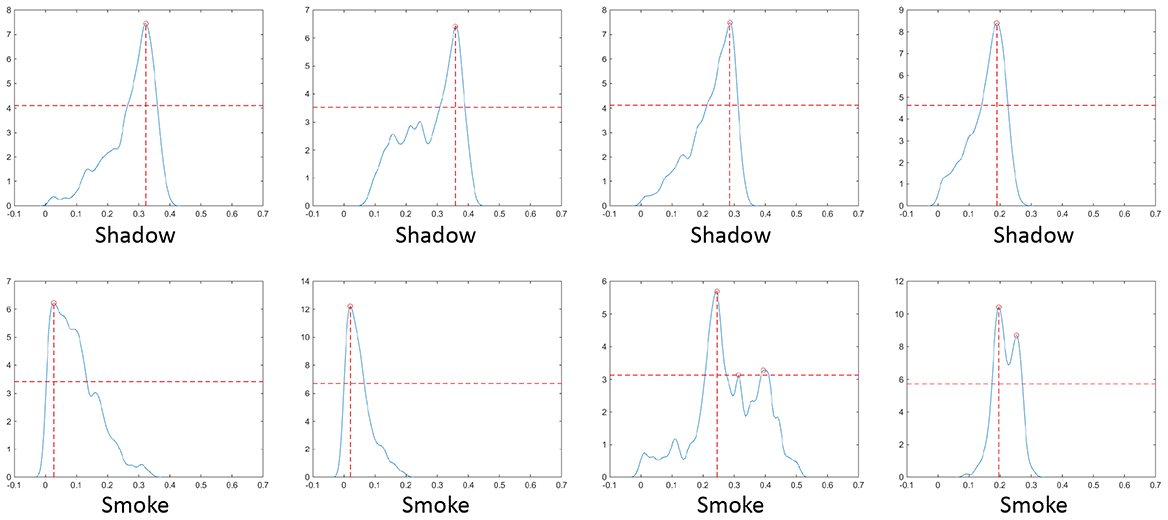}
	\caption{Each small graph shows the probability density function of a smoke or shadow region's corresponding pixel values in $S_{t}$ (see Figure~\ref{fig:TEX}) using kernel density estimation. The x-axis represents the pixel values in $S_{t}$. The horizontal red line is the threshold for computing number of peaks. The vertical red line indicates the pixel value of the highest peak.}
	\label{fig:shadow-detection}
\end{figure}

Finally we remove regions which may contain shadow. The algorithm performs background subtraction using (\ref{eq:BS}) on $I_{t}$ and $B_{t}$ to obtain $S_{t}=\bgSub(I_{t},B_{t})$ in Figure~\ref{fig:TEX}. Then we compute the probability density function (PDF) of each region's corresponding pixel values in $S_{t}$ using kernel density estimation~\cite{Silverman1986} with a Gaussian kernel.
\begin{equation}
\!\!\!\!\!\!\!\! \hat{p}(x) = \frac{1}{n}\sum_{i=1}^{n}\frac{1}{h}K\bigg(\frac{x-X_{i}}{h}\bigg) \;\;\mathrm{where}\;\; X_{i} \in S_{t}
\end{equation}
Because the PDF of shadow and smoke regions have distinct characteristics (see Figure~\ref{fig:shadow-detection}), we can describe shadow regions by utilizing (\ref{eq:shadow})
\begin{equation}\label{eq:shadow}
\!\!\!\!\!\!\!\! \argmax_{x}{\;p(x)}>t_{8} \;\;\;\mathrm{and}\;\; \sum_{x_{i} \in X}{\textbf{1}_{\{p(x_{i})>t_{9}\}}}<t_{10}
\end{equation}
where $x$ indicates pixel values, $p(x)$ is the probability density function, $\argmax_{x}{\;p(x)}$ means the pixel value of the highest peak, $X$ is a set of pixel values of the corresponding peaks, $\textbf{1}_{A}$ is the indicator function of a set $A$, and $\sum_{x_{i} \in X}{\textbf{1}_{\{p(x_{i})>t_{9}\}}}$ is the number of peaks having their heights exceed a certain threshold.

Applying all the above region filtering steps on $R_{smooth}$ yields $R_{filter}$ (see Figure~\ref{fig:TEX}). We compute a mask $M_{t}$ which is a binary image based on $R_{filter}$ and output the response at time $t$ as the sum of all pixel values in mask $M_{t}$.

\subsection{Event Detection}

\begin{figure}[t]
	\centering
	\includegraphics[width=1\columnwidth]{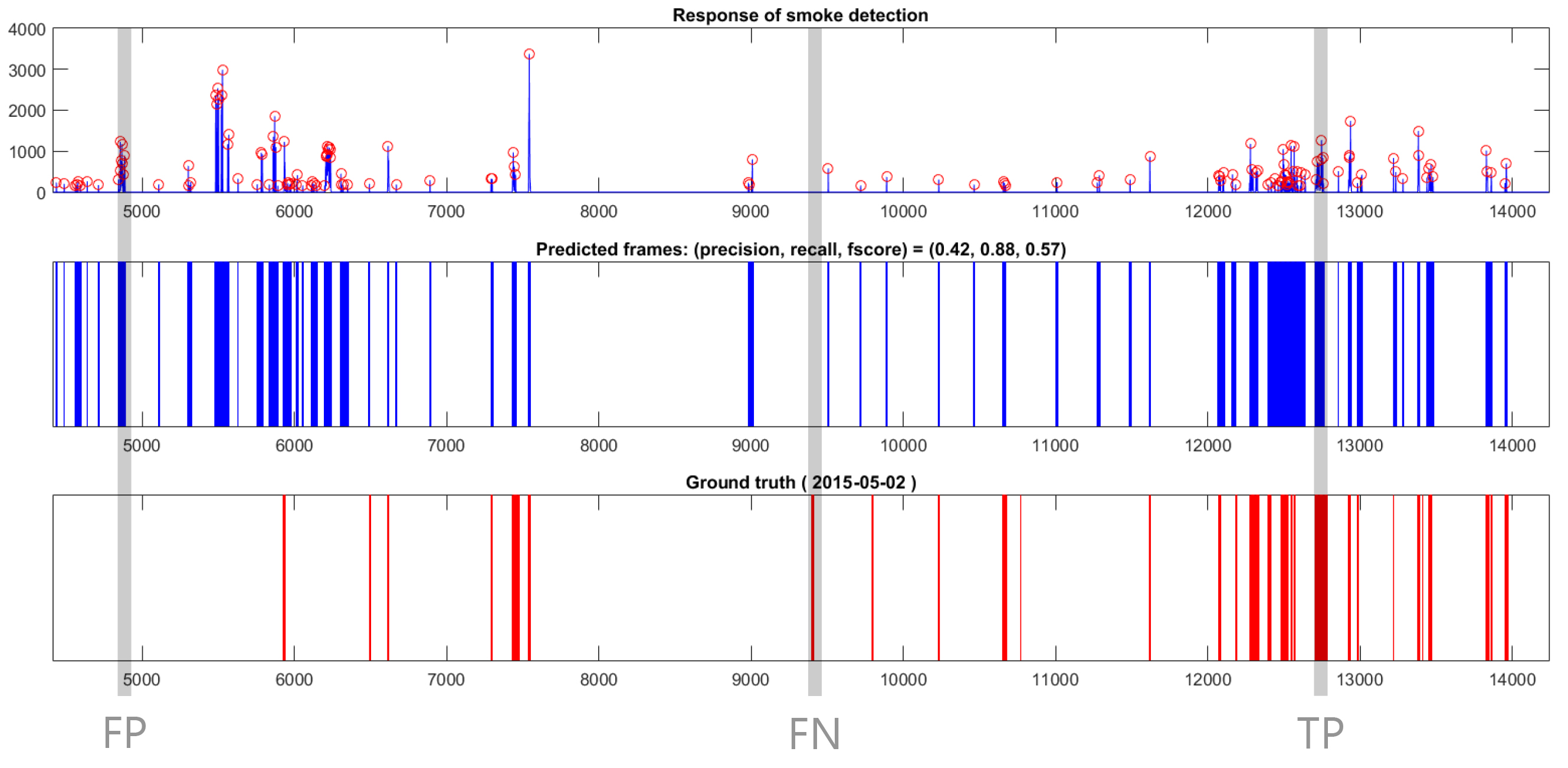}
	\caption{This figure shows the result of smoke detection. The x-axis and y-axis indicate the frame number and the amount of pixels identified as smoke. The bottom graph is the ground truth of May 2, 2015. The top and middle graphs show the response and the prediction of all daytime frames. The red circles in the top graph represent the local peaks. The gray bars indicate true positive (TP), false positive (FP), and false negative (FN).}
	\label{fig:result}
\end{figure}

Event Detection identifies the starting and ending time of fugitive emissions. We first select daytime frames for each day and ignore nighttime ones because of lighting issues. Next, we apply change detection, texture segmentation, and region filtering on these frames to obtain a time-series signal (see the top chart in Figure~\ref{fig:result}). Each value in the time-series signal represents the number of smoke pixels in a corresponding video frame. Then we compute segments in the time-series signal by finding peaks and corresponding peak widths. Finally we merge nearby segments into events (see the middle chart in Figure~\ref{fig:result}).

\section{Experiment}

\begin{table}
	\caption{The evaluation of all daytime frames for 9 days on May 2015.}
	\begin{center}\label{tb:fscore_all}
		\begin{tabular}{|l|l|l|l|l|l|l|}
			\hline Date & TP & FP & FN & Precision & Recall & F-score \\
			\hline May 1 & 15 & 36 & 4 & 0.2941 & 0.7895 & 0.4286 \\
			\hline May 2 & 21 & 29 & 3 & 0.4200 & 0.8750 & 0.5676 \\
			\hline May 3 & 24 & 28 & 8 & 0.4615 & 0.7500 & 0.5714 \\
			\hline May 4 & 25 & 25 & 5 & 0.5000 & 0.8333 & 0.6250 \\
			\hline May 5 & 14 & 19 & 4 & 0.4242 & 0.7778 & 0.5490 \\
			\hline May 6 & 17 & 11 & 4 & 0.6071 & 0.8095 & 0.6939 \\
			\hline May 7 & 26 & 16 & 3 & 0.6190 & 0.8966 & 0.7324 \\
			\hline May 8 & 22 & 22 & 4 & 0.5000 & 0.8462 & 0.6286 \\
			\hline May 9 & 16 & 23 & 1 & 0.4103 & 0.9412 & 0.5714 \\
			\hline Avg & & & & 0.4707 & 0.8355 & 0.5964 \\
			\hline 
		\end{tabular} 
	\end{center}
\end{table}

\begin{table}
	\caption{The evaluation of all daytime frames (exclude frames containing steam) for 9 days on May 2015.}
	\begin{center}\label{tb:fscore_all_no_steam}
		\begin{tabular}{|l|l|l|l|l|l|l|}
			\hline Date &  TP & FP & FN & Precision & Recall & F-score \\
			\hline May 1 & 13 & 8 & 4 & 0.6190 & 0.7647 & 0.6842 \\
			\hline May 2 & 18 & 11 & 3 & 0.6207 & 0.8571 & 0.7200 \\
			\hline May 3 & 24 & 19 & 6 & 0.5581 & 0.8000 & 0.6575 \\
			\hline May 4 & 25 & 17 & 4 & 0.5952 & 0.8621 & 0.7042 \\
			\hline May 5 & 13 & 9 & 3 & 0.5909 & 0.8125 & 0.6842 \\
			\hline May 6 & 15 & 4 & 4 & 0.7895 & 0.7895 & 0.7895 \\
			\hline May 7 & 26 & 6 & 3 & 0.8125 & 0.8966 & 0.8525 \\
			\hline May 8 & 22 & 18 & 4 & 0.5500 & 0.8462 & 0.6667 \\
			\hline May 9 & 14 & 17 & 1 & 0.4516 & 0.9333 & 0.6087 \\
			\hline Avg & & & & 0.6209 & 0.8402 & 0.7075 \\
			\hline 
		\end{tabular} 
	\end{center}
\end{table}

\begin{table}
	\caption{Evaluation of the smoke detection algorithm on 12 randomly chosen days for each month in 2015. TP, FP, and FN indicates true positive, false positive, and false negative respectively.}
	\begin{center}\label{tb:fscore}
		\begin{tabular}{|l|c|c|c|c|c|c|}
			\hline Date\;\;\;\;\; & TP & FP & FN & Precision & Recall & F-score \\
			\rowcolor{gray} \hline Dec 22 & 18 & 21 & 7 & 0.4615 & 0.7200 & 0.5625 \\
			\rowcolor{gray} \hline Nov 15 & 18 & 6 & 1 & 0.7500 & 0.9474 & 0.8372 \\
			\rowcolor{gray} \hline Oct 05 & 27 & 23 & 0 & 0.5400 & 0.9643 & 0.6923 \\
			\rowcolor{gray} \hline Sep 09 & 10 & 35 & 8 & 0.2222 & 0.5556 & 0.3175 \\
			\rowcolor{gray} \hline Aug 13 & 28 & 35 & 2 & 0.4444 & 0.9333 & 0.6022 \\
			\rowcolor{gray} \hline Jul 08 & 15 & 35 & 9 & 0.3000 & 0.6250 & 0.4054 \\
			\rowcolor{gray} \hline Jun 11 & 22 & 14 & 4 & 0.6111 & 0.8462 & 0.7097 \\
			\rowcolor{gray} \hline May 28 & 24 & 17 & 3 & 0.5854 & 0.8889 & 0.7059 \\
			\rowcolor{gray} \hline Apr 02 & 15 & 28 & 10 & 0.3488 & 0.6000 & 0.4412 \\
			\hline Mar 06 & 1 & 8 & 15 & 0.1111 & 0.0625 & 0.0800 \\
			\hline Feb 10 & 3 & 32 & 10 & 0.0857 & 0.2308 & 0.1250 \\
			\hline Jan 26 & 1 & 5 & 2 & 0.1667 & 0.3333 & 0.2222 \\
			\hline 
		\end{tabular} 
	\end{center}
\end{table}

\begin{figure}
	\centering
	\includegraphics[width=1\columnwidth]{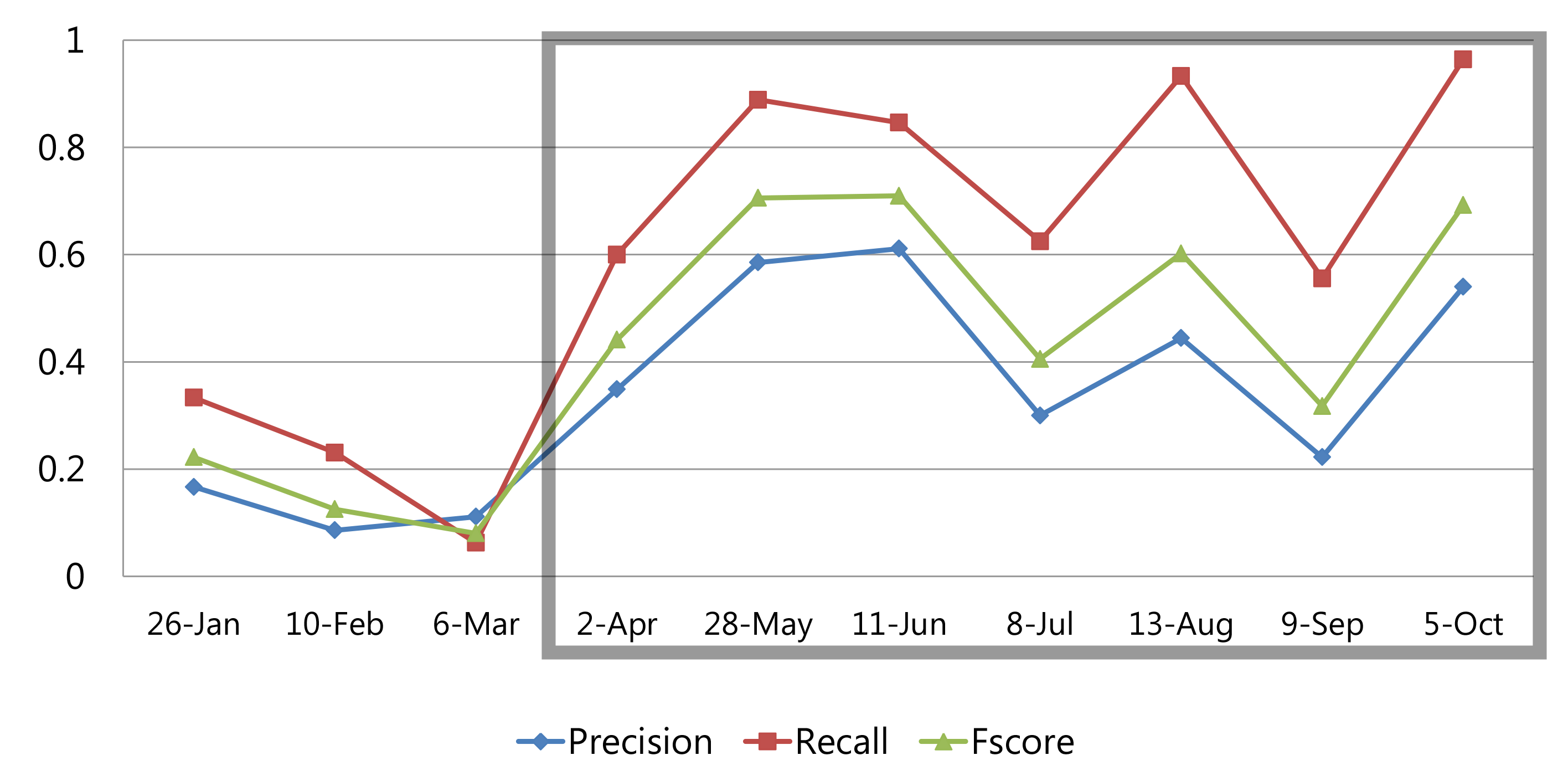}
	\caption{Evaluation of the smoke detection algorithm on 12 randomly chosen days for each month in 2015.}
	\label{fig:evaluation-date}
\end{figure}

We used MATLAB to develop the smoke detection algorithm and VLFeat~\cite{vedaldi08vlfeat} library to run the accelerated k-means++ algorithm for clustering feature vectors during the texture segmentation step. Each timelapse of a day consisted of 16838 frames. We ran the smoke detection algorithm on a window with 496-by-528 pixels in the timelapse video for 21 days in 2015 during daytime. The processing time was 30 minutes on average for a day by using all cores on a workstation with two hex-core CPU (Intel Xeon X5670).

We manually labeled these 21 days to evaluate the performance of the algorithm. The bottom graph of Figure~\ref{fig:result} shows the ground truth labels on May 2. The middle and bottom graphs demonstrate the response and the prediction of smoke emissions. Table~\ref{tb:fscore_all} and \ref{tb:fscore_all_no_steam} show the results of evaluation from May 1st to 9th with and without the frames having steam. Table~\ref{tb:fscore} shows the accuracy of twelve randomly picked days for each month in 2015.

We calculate true positives (TP), false positives (FP), false negatives (FN). Denote the boolean array of ground truth labels $G$ and predictions $P$ which contains only true and false entries. We first group the continuous true entries in $G$ into a series of segments and apply the same process on prediction $P$. Next, for each segment in $P$, denote the starting and ending frame indices $m_{p}$ and $n_{p}$. We mark a segment as a true positive if 30\% of the entries in the segment contains true ground truth labels, which is described in (\ref{eq:TP}). Otherwise, we mark the segment as a false positive.
\begin{equation}\label{eq:TP}
\frac{\sum_{i=m_{p}}^{n_{p}}{G(i)}}{n_{p}-m_{p}+1}>0.3
\end{equation}
For each segment in $G$, denote the starting and ending frame indices $m_{g}$ and $n_{g}$. We mark a segment as a false negative if $\sum_{i=m_{g}}^{n_{g}}{P(i)} = 0$, which means no entries in the segment contains true predictions. Finally, we compute precision (PR), recall (RE), and F-score by using (\ref{eq:fscore}).
\begin{equation}\label{eq:fscore}
\begin{array}{lll}
\text{PR} & = & \text{TP}/(\text{TP}+\text{FP}) \\
\text{RE} & = & \text{TP}/(\text{TP}+\text{FN}) \\
\text{F-score} & = & 2*\text{PR}*\text{RE}/(\text{PR}+\text{RE})
\end{array}
\end{equation}

\section{Visualization}

\begin{figure}
	\centering
	\includegraphics[width=1\columnwidth]{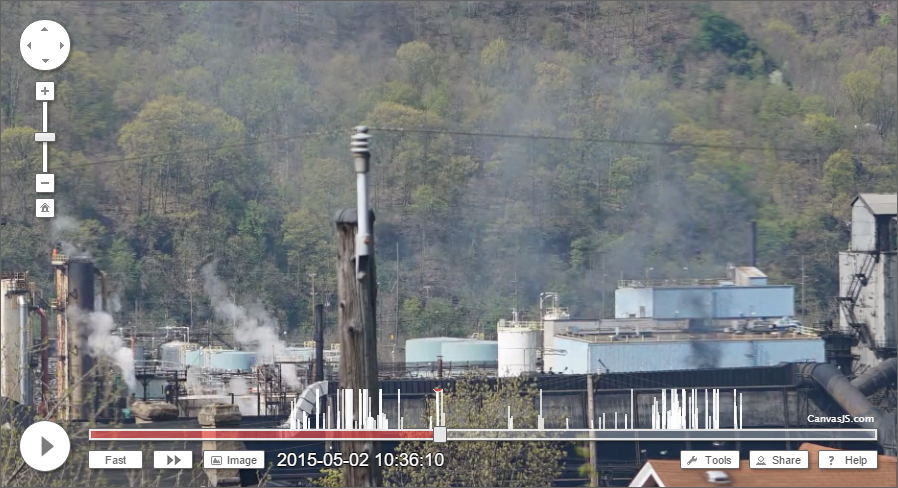}
	\caption{This figure shows the timelapse viewer with an interactive timeline which visualizes the results of smoke detection and provides indicators for seeking to interesting events. The fast-forwarding button only plays interesting segments and skips the rest. The image button gives a collection of animated images (see Figure~\ref{fig:collection}) which are likely to contain smoke.}
	\label{fig:UI}
\end{figure}

\begin{figure}
	\centering
	\includegraphics[width=1\columnwidth]{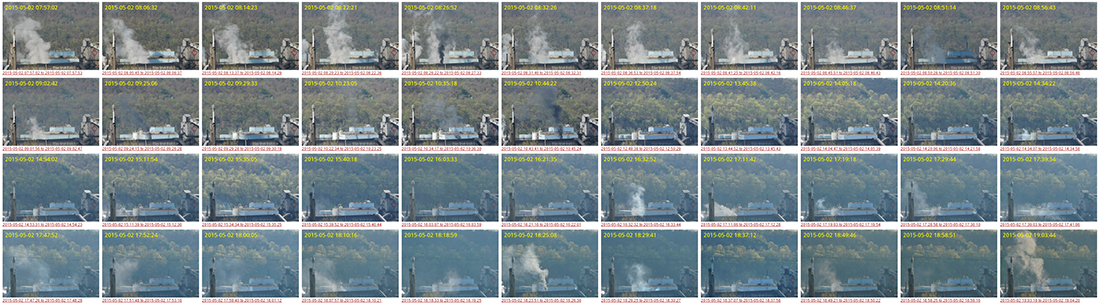}
	\caption{This figure shows a part of the collection of animated images generated by the timelapse viewer according to the results of smoke detection on May 2, 2015. The local community can select desired images and drag them into a Google Doc for documentation, presentation, and storytelling. The hyperlinks under the animated images redirect users back to the timelapse viewer upon mouse clicking.}
	\label{fig:collection}
\end{figure}

We apply the smoke detection algorithm on the nine months in 2015 with better accuracies (see the highlighted rows in Table~\ref{tb:fscore} and Figure \ref{fig:evaluation-date}). We visualize the results by using an interactive timeline, an autonomous fast-forwarding function, and a visual summary (see Figure~\ref{fig:UI} and Figure~\ref{fig:collection}). The programing languages of the visualization are JavaScript and HTML. The interactive timeline serves as an indicator; community members can click on a spike on the timeline to seek to a frame with fugitive emissions. The fast-forwarding function plays only the smoke-sensed time segments in the timelapse video and skips the rest. The visual summary provides a collection of animated images generated from the time segments. Each image has a hyperlink to the timelapse viewer at a particular time and view. Community members can share or collect desired smoke images in online documents as visual evidence.


\chapter{Discussion and Future Work}

The system significantly reduces the workload and the time of searching and documenting fugitive emissions so that local community members can focus on the content of documentation instead of laborious and time-consuming works. It takes approximately 840 seconds to play the timelapse video at 12 frames per second for a day. With the fast-forwarding feature, the local community members can now browse candidate frames containing smoke for a whole day in a shorter time. Moreover, manually searching and generating animated images of smoke emissions through all daytime frames takes about 1.5 hours. The viewer can now provide a collection of images as shown in Figure~\ref{fig:collection} almost instantly when citizen scientists click the image button on the user interface. 

The system has several limitations. First, the algorithm uses a heuristic approach. There are numerous thresholds that one needs to adjust for the algorithm. We use only one set of tuning parameters. It is an open question as to whether the same parameter set works for daytime frames in other days. In addition, since the lighting conditions of nighttime and daytime are different, detecting smoke emissions at nighttime may require another algorithm.

Second, the system produces a large portion of false positives caused by steam or fast-moving shadows and a small portion of false negatives caused by low opacity or small-sized smoke. The algorithm has difficulties in completely excluding fast-moving shadows due to the combination of wind and partly cloudy conditions. It also cannot distinguish between white smoke and steam clearly. These are the reasons why the precisions in table~\ref{tb:fscore_all} are low. The overall accuracy increases if we remove the frames having steam (see Table~\ref{tb:fscore_all_no_steam}). We use frame differencing (see section~\ref{sec:IICD}) to remove steam regions, which assumes three conditions: (1) the size of steam is large; (2) wind blows at a constant direction slowly; and (3) steam emits from places that are inside the detection window. Nevertheless, steam may have low opacity or change a lot across multiple frames due to high wind speed, which makes it numerically similar to white smoke (see Figure~\ref{fig:smoke-type-2}). Currently we rely on citizen scientists to exclude steam using human vision from the presented collection of animated images.

In the future, we plan to generalize this work by obtaining more labels using crowdsourcing, turning smoke images and these labels into a reusable dataset, and training a classifier using these labels.

\backmatter


\renewcommand{\bibsection}{\chapter{\bibname}}
\bibliographystyle{plainnat}
\bibliography{ref/ref}

\begin{thebibliography}{26}
\providecommand{\natexlab}[1]{#1}
\providecommand{\url}[1]{\texttt{#1}}
\expandafter\ifx\csname urlstyle\endcsname\relax
  \providecommand{\doi}[1]{doi: #1}\else
  \providecommand{\doi}{doi: \begingroup \urlstyle{rm}\Url}\fi

\bibitem[TMv()]{TMviewer}
{Timelapse Viewer}.
\newblock \url{http://timemachine.cmucreatelab.org/}.

\bibitem[Arthur and Vassilvitskii(2007)]{Arthur2007}
David Arthur and Sergei Vassilvitskii.
\newblock K-means++: The advantages of careful seeding.
\newblock In \emph{Proceedings of the Eighteenth Annual ACM-SIAM Symposium on
  Discrete Algorithms}, pages 1027--1035, Philadelphia, PA, USA, 2007. Society
  for Industrial and Applied Mathematics.

\bibitem[Calderara et~al.(2008)Calderara, Piccinini, and
  Cucchiara]{Calderara2008}
Simone Calderara, Paolo Piccinini, and Rita Cucchiara.
\newblock Smoke detection in video surveillance: A mog model in the wavelet
  domain.
\newblock In \emph{Computer Vision Systems}, volume 5008 of \emph{Lecture Notes
  in Computer Science}, pages 119--128. Springer Berlin Heidelberg, 2008.

\bibitem[\c{C}elik et~al.(2007)\c{C}elik, \"{O}zkaramanli, and
  Demirel]{Celik2007}
Turgay \c{C}elik, H\"{u}seyin \"{O}zkaramanli, and Hasan Demirel.
\newblock {Fire and smoke detection without sensors: Image processing based
  approach}.
\newblock In \emph{European Signal Processing Conference}, pages 1794--1798,
  2007.

\bibitem[Cheung and Kamath(2005)]{Cheung2005}
Sen-Ching~S. Cheung and Chandrika Kamath.
\newblock Robust background subtraction with foreground validation for urban
  traffic video.
\newblock \emph{EURASIP J. Appl. Signal Process.}, 2005:\penalty0 2330--2340,
  January 2005.

\bibitem[Collins et~al.(2000)Collins, Lipton, Kanade, Fujiyoshi, Duggins, Tsin,
  Tolliver, Enomoto, and Hasegawa]{Collins2000}
Robert Collins, Alan Lipton, Takeo Kanade, Hironobu Fujiyoshi, David Duggins,
  Yanghai Tsin, David Tolliver, Nobuyoshi Enomoto, and Osamu Hasegawa.
\newblock A system for video surveillance and monitoring.
\newblock Technical report, The Robotics Institute, Carnegie Mellon University,
  Pittsburgh, PA, May 2000.

\bibitem[Elkan(2003)]{Elkan2003}
Charles Elkan.
\newblock {Using the Triangle Inequality to Accelerate K-Means}.
\newblock In \emph{Proceedings of the Twentieth International Conference on
  Machine Learning (ICML-2003)}, 2003.

\bibitem[Friedman and Russell(1997)]{Friedman1997}
Nir Friedman and Stuart Russell.
\newblock Image segmentation in video sequences: A probabilistic approach.
\newblock In \emph{Proceedings of the Thirteenth Conference on Uncertainty in
  Artificial Intelligence}, UAI'97, pages 175--181, San Francisco, CA, USA,
  1997. Morgan Kaufmann Publishers Inc.

\bibitem[Gubbi et~al.(2009)Gubbi, Marusic, and Palaniswami]{Gubbi2009}
Jayavardhana Gubbi, Slaven Marusic, and Marimuthu Palaniswami.
\newblock Smoke detection in video using wavelets and support vector machines.
\newblock \emph{Fire Safety Journal}, 44:\penalty0 1110 -- 1115, 2009.

\bibitem[Haklay(2013)]{Haklay2013}
Muki Haklay.
\newblock Citizen science and volunteered geographic information: Overview and
  typology of participation.
\newblock In \emph{Crowdsourcing Geographic Knowledge}, pages 105--122.
  Springer Netherlands, 2013.

\bibitem[Hohberg(2015)]{Hohberg2015}
Simon~Philipp Hohberg.
\newblock Wildfire smoke detection using convolutional neural networks.
\newblock Technical report, Freie Universität Berlin, Berlin, Germany,
  September 2015.

\bibitem[Kopilovic et~al.(2000)Kopilovic, Vagvolgyi, and
  Sziranyi]{Kopilovic2000}
I.~Kopilovic, B.~Vagvolgyi, and T.~Sziranyi.
\newblock Application of panoramic annular lens for motion analysis tasks:
  surveillance and smoke detection.
\newblock In \emph{Pattern Recognition, 2000. Proceedings. 15th International
  Conference on}, volume~4, pages 714--717 vol.4, 2000.

\bibitem[Kosara and MacKinlay(2013)]{Kosara2013}
Robert Kosara and Jock MacKinlay.
\newblock Storytelling: The next step for visualization.
\newblock \emph{Computer}, 46\penalty0 (5):\penalty0 44--50, 2013.

\bibitem[Laws(1980)]{Laws1980}
Kenneth~I. Laws.
\newblock \emph{Textured Image Segmentation}.
\newblock PhD thesis, University of Southern California, Los Angeles., Jan
  1980.

\bibitem[Lee et~al.(2012)Lee, Lin, Hong, and Su]{Lee2012}
Chen-Yu Lee, Chin-Teng Lin, Chao-Ting Hong, and Miin-Tsair Su.
\newblock {Smoke Detection Using Spatial and Temporal Analysis}.
\newblock \emph{International Journal of Innovative Computing, Information and
  Control}, 8:\penalty0 4749--4770, 2012.

\bibitem[Ma et~al.(2012)Ma, Liao, Frazier, Hauser, and Kostis]{Ma2012}
Kwan~Liu Ma, Isaac Liao, Jennifer Frazier, Helwig Hauser, and Helen-Nicole
  Kostis.
\newblock Scientific storytelling using visualization.
\newblock \emph{Computer Graphics and Applications, IEEE}, 32\penalty0
  (4):\penalty0 12--19, 2012.

\bibitem[Malik et~al.(2001)Malik, Belongie, Leung, and Shi]{Malik2001}
Jitendra Malik, Serge Belongie, Thomas Leung, and Jianbo Shi.
\newblock Contour and texture analysis for image segmentation.
\newblock \emph{International Journal of Computer Vision}, 43:\penalty0 7--27,
  2001.

\bibitem[Radke et~al.(2005)Radke, Andra, Al-Kofahi, and Roysam]{Radke2005}
R.J. Radke, S.~Andra, O.~Al-Kofahi, and B.~Roysam.
\newblock Image change detection algorithms: a systematic survey.
\newblock \emph{Image Processing, IEEE Transactions on}, 14\penalty0
  (3):\penalty0 294--307, March 2005.

\bibitem[Sargent et~al.(2010)Sargent, Bartley, Dille, Keller, and
  Nourbakhsh]{Sargent2010}
Randy Sargent, Chris Bartley, Paul Dille, Jeff Keller, and Illah Nourbakhsh.
\newblock {Timelapse GigaPan: Capturing, Sharing, and Exploring Timelapse
  Gigapixel Imagery}.
\newblock In \emph{Fine International Conference on Gigapixel Imaging for
  Science}, 2010.

\bibitem[Silverman(1986)]{Silverman1986}
Bernard~Walter Silverman.
\newblock Density estimation for statistics and data analysis.
\newblock In \emph{Computer Vision Systems}, Number 26 in Monographs on
  Statistics and Applied Probability. Chapman \& Hall, 1986.

\bibitem[Silvertown(2009)]{Silvertown2009}
Jonathan Silvertown.
\newblock A new dawn for citizen science.
\newblock \emph{Trends in Ecology \& Evolution}, 24\penalty0 (9):\penalty0
  467--471, 2009.

\bibitem[Stauffer and Grimson(1999)]{Stauffer1999}
Chris Stauffer and W.E.L. Grimson.
\newblock Adaptive background mixture models for real-time tracking.
\newblock In \emph{IEEE Computer Society Conference on Computer Vision and
  Pattern Recognition}, volume~2, 1999.

\bibitem[Tian et~al.(2015)Tian, Li, Ogunbona, Wang, Reid, Saito, and
  Yang]{Tian2015}
Hongda Tian, Wanqing Li, Philip Ogunbona, editor="Cremers~Daniel Wang, Lei",
  Ian Reid, Hideo Saito, and Ming-Hsuan Yang.
\newblock \emph{Single Image Smoke Detection}, chapter Computer Vision -- ACCV
  2014: 12th Asian Conference on Computer Vision, Singapore, Singapore,
  November 1-5, 2014, Revised Selected Papers, Part II, pages {87--101}.
\newblock Springer International Publishing, 2015.

\bibitem[Toreyin et~al.(2005)Toreyin, Dedeoglu, and Cetin]{Toreyin2005}
B.U. Toreyin, Y.~Dedeoglu, and A.E. Cetin.
\newblock Wavelet based real-time smoke detection in video.
\newblock In \emph{Signal Processing Conference, 2005 13th European}, pages
  1--4, Sept 2005.

\bibitem[Vedaldi and Fulkerson(2008)]{vedaldi08vlfeat}
A.~Vedaldi and B.~Fulkerson.
\newblock {VLFeat}: An open and portable library of computer vision algorithms,
  2008.
\newblock \url{http://www.vlfeat.org/}.

\bibitem[Zuiderveld(1994)]{Zuiderveld1994}
Karel Zuiderveld.
\newblock {Contrast Limited Adaptive Histograph Equalization}.
\newblock pages 474--485. Academic Press Professional, Inc., 1994.

\end{thebibliography}

\end{document}